\documentclass[letterpaper]{article} 
\usepackage{aaai24}  
\usepackage{times}  
\usepackage{helvet}  
\usepackage{courier}  
\usepackage[hyphens]{url}  
\usepackage{graphicx} 
\usepackage{natbib}  
\usepackage{caption} 
\usepackage{algorithm}
\usepackage{algorithmic}

\usepackage{newfloat}
\usepackage{listings}
\DeclareCaptionStyle{ruled}{labelfont=normalfont,labelsep=colon,strut=off} 
\floatstyle{ruled}
\newfloat{listing}{tb}{lst}{}
\floatname{listing}{Listing}
\usepackage{booktabs}
\newcommand{\ra}[1]{\renewcommand{\arraystretch}{#1}}
\usepackage{multirow}
\usepackage{amsfonts} 
\usepackage{amsmath}
\usepackage{xcolor}
\usepackage{pifont}

\title{LADDER: An Efficient Framework for Video Frame Interpolation}
\author{
    \small Tong Shen \textsuperscript{\rm 1},
            Dong Li \textsuperscript{\rm 1},
            Ziheng Gao \textsuperscript{\rm 1},
            Lu Tian \textsuperscript{\rm 1},
            Emad Barsoum \textsuperscript{\rm 1}
}
\affiliations{
    \textsuperscript{\rm 1} Advanced Micro Devices, USA
    
    \small \{t.shen, d.li, ziheng.gao, lu.tian, ebarsoum\}amd.com
}

\usepackage{bibentry}

\begin{document}

\maketitle

\begin{abstract}
Video Frame Interpolation (VFI) is a crucial technique in various applications such as slow-motion generation, frame rate conversion, video frame restoration etc. This paper introduces an efficient video frame interpolation framework that aims to strike a favorable balance between efficiency and quality. Our framework follows a general paradigm consisting of a flow estimator and a refinement module, while incorporating carefully designed components. First of all, we adopt depth-wise convolution with large kernels in the flow estimator that simultaneously reduces the parameters and enhances the receptive field for encoding rich context and handling complex motion. Secondly, diverging from a common design for the refinement module with a UNet-structure (encoder-decoder structure), which we find redundant, our decoder-only refinement module directly enhances the result from coarse to fine features, offering a more efficient process. In addition, to address the challenge of handling high-definition frames, we also introduce an innovative HD-aware augmentation strategy during training, leading to consistent enhancement on HD images. Extensive experiments are conducted on diverse datasets, Vimeo90K, UCF101, Xiph and SNU-FILM. The results demonstrate that our approach achieves state-of-the-art performance with clear improvement while requiring much less FLOPs and parameters, reaching to a better spot for balancing efficiency and quality.
\end{abstract}

\section{Introduction}

\begin{figure}[t]
\centering
\includegraphics[width=1\columnwidth]{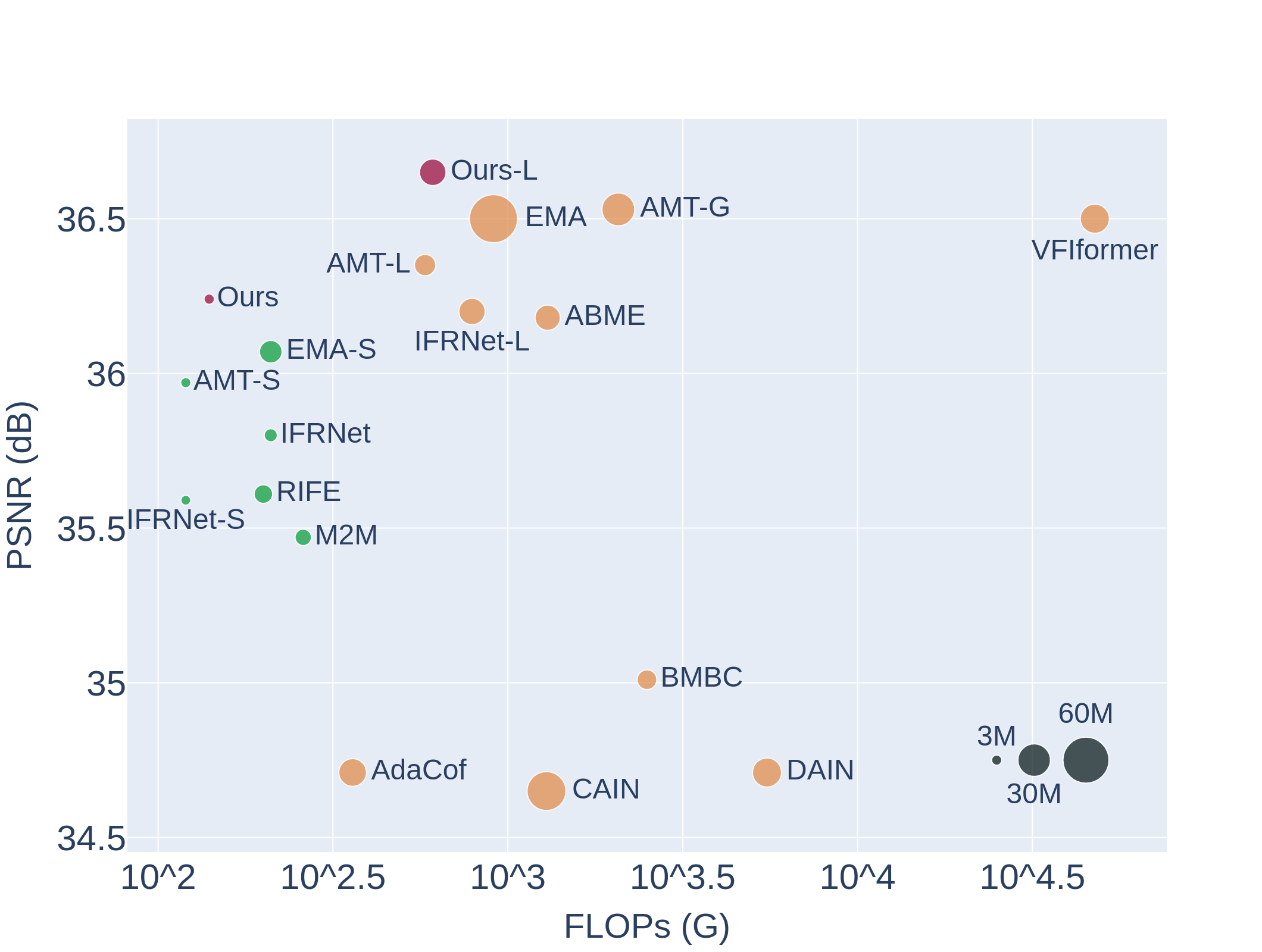} 
\caption{Model comparison on Vimeo90K dataset in terms of FLOPs, PSNR and number of parameters. We roughly divide the models into light-weight and large models, which are shown with green color and orange color respectively. The marker size indicates the number of parameters ranging from 1.4M to 60+M. We present two versions of our method, which are colored red. For both light-weight and large groups, we achieve higher PSNR with much lower complexity, reaching to a better balancing spot.}
\label{model_comparison}
\end{figure}

Video Frame Interpolation (VFI) is an important task in video processing aiming to synthesize intermediate frames between existing frames. VFI can support many downstream tasks such as slow-motion generation \cite{superslomo}, video compression \cite{wu2018vcii}, novel-view synthesis \cite{zhou2016view} and video prediction \cite{yue2022videopred}.



Various approaches have been proposed to tackle the problem, but it is still challenging to find an optimal balance spot between efficiency and quality. For example, VFIformer \cite{vfiformer}, a previous state-of-the-art method, achieves PSNR of 36.5 dB on Vimeo90K \cite{vimeo} dataset and outperforms another light-weight method, RIFE \cite{rife}, by a large margin, which obtains PSNR of 35.61 dB. However, to compare the FLOPs of these two models, VFIformer requires nearly 238 times more FLOPs than RIFE for the same inputs. Obviously, it is still worth exploring a better trade-off between efficiency and quality. To this end, we propose an efficient VFI framework with carefully designed components that achieves new state-of-the-art performance with clear improvement while requiring much less FLOPS and parameters (as shown in Figure \ref{model_comparison}). 

A widely used form for flow-based methods mainly comprises three parts, a feature extractor, a flow estimator and a refinement module. Our general design philosophy lies in these components:

1) For the feature extractor, since good features not only provide good descriptors for flow estimation but also high-quality evidence for synthesizing frames, we adopt an effective feature extraction process inspired from \cite{ema} that has convolution blocks for low-level features and light-transformer blocks for high-level features. Different from the original structure, we do not extract motion features for simplicity.

2) For the flow estimator, this module is typically devised to operate in a coarse-to-fine fashion, progressively decoding and refining the flow estimation from an initial coarse approximation to the final output. 
A key motivation here is that flow estimation usually requires a large field of view (FoV) to capture large motions and rich context. Stacking up multiple $3 \times 3$ convolutions might be sufficient for low resolution feature maps, where one spatial feature vector corresponds to a large region in pixel space thanks to big down-sampling scales. However, it might not suffice for high resolution feature maps. Certainly, increasing kernel size or stacking more layers can further enlarge FoV, but it comes at the expense of largely increasing computational cost. Therefore, we propose to use large-kernel depth-wise convolution for those high resolution features. The idea of using large kernels has been investigated for classification, segmentation and object detection \cite{largekernel, replknet} but rarely studied in VFI, which motivates us to explore the effectiveness. Our carefully designed flow estimator actually consists of two types of encoders: \textbf{Low-res encoder} with normal $3\times3$ convolution for low resolution features and \textbf{High-res encoder} with large-kernel depth-wise convolution for high resolution features.

 3) For the refinement module, existing approaches \cite{ema, vfiformer, rife} commonly adopt a UNet structure \cite{unet}, featuring a encoder and a decoder.
 Our key observation is that the feature extractor has already encoded the images into high-quality features on different levels, which is already playing the role of the encoder, and it is not necessary to have another encoder inside of the refinement module. Therefore, diverging from the UNet-like design, we propose to use a decoder-only structure with only three levels that shares the computation and features with the feature extractor and directly estimates the final prediction from previously calculated results, which enables efficient and effective refinement process.

 4) Additionally, we notice that our model trained on Standard-Definition (SD) images (e.g. Vimeo90K \cite{vimeo}) does not always give consistent increase of performance on High-Definition (HD) images (e.g. Xiph \cite{xiph} and SNU-FILM \cite{cain}) because of the large size of the image and the limited motion capture ability of the network. To address this issue, we also design a HD-aware augmentation strategy that enables the model's ability to harness both the down-scaled flow and original flow, which leads to more stable enhancement in handling diverse image resolutions.

Combining above designs, we propose LADDER, a VFI framework with \textbf{LA}rge-kernel \textbf{D}epth-wise Convolution and \textbf{DE}coder-only \textbf{R}efinement.
Our method can be trained solely with image triplets without requiring extra supervisions or any pre-trained network \cite{abme, dain, softsplat}.

 To sum up, our contributions are:
 \begin{itemize}
 \item We propose LADDER for VFI task with a carefully designed flow estimator and refinement module. For the flow estimator, we leverage depth-wise convolution with large kernels for reduced computational overhead and large field of view. The refinement module takes the form of a decoder-only structure, featuring efficiency and effectiveness.

 \item We also design a HD-aware augmentation strategy to address the "performance drop" issue on HD images.
 \item Extensive experiments are conducted on various datasets and our light-weight model, alongside with its more extensive counterpart, achieves state-of-the-art performance with clear improvement while demanding much less FLOPs and parameters.
\end{itemize}
\section{Related Work}

\textbf{Video Frame Interpolation} VFI methods can be roughly categorized into two groups. \textit{Kernel-based methods} represent motion implicitly by using kernels, either a regular \cite{adapconv, sepconv, imnet} or deformable kernel \cite{adacof, vfidsc}. The intermediate frame is generated by applying local convolution on the input frames. \textit{Flow-based methods} formulate motion by optical flow, which is a common representation to describe motion between frames. Some approaches \cite{cavfi, qvi_nips19} rely on a pre-trained optical flow model for off-the-shelf flow estimation and further predict the intermediate frame by a synthesis network. After that, the concept of task-oriented flow is proposed \cite{toflow}, defining a flow representation tailored for VFI task. Some methods directly predict intermediate flows instead of approximating them \cite{bmbc, abme, rife}. With the increasing popularity of Transformer, there are works \cite{ema, vfiformer, amt} incorporating Transformer structures into VFI models to model long-range correlation and capture larger motions.

\textbf{Light-weight VFI.} Some approaches aim at designing light-weight models for VFI task, which is also one of our goals in the work. \cite{rife} propose an efficient VIF pipeline termed IFNet that consists of multiple IFBlocks to estimate flows at different levels. Each block is designed for direct intermediate flow estimation, enabling both effectiveness and simplicity. A privileged distillation scheme is also introduced to further improve the performance. \cite{ifrnet} present an efficient encoder-decoder based network, named IFRNet, for fast VFI. It employs a decoder to jointly refine the intermediate flow and generate a intermediate feature to produce the final synthesized frame. \cite{ema} propose an efficient transformer-based network that leverages both inter-frame appearance features and motion features for VFI.

\section{Method}

\begin{figure*}[t]
\centering
\includegraphics[width=0.93\textwidth]{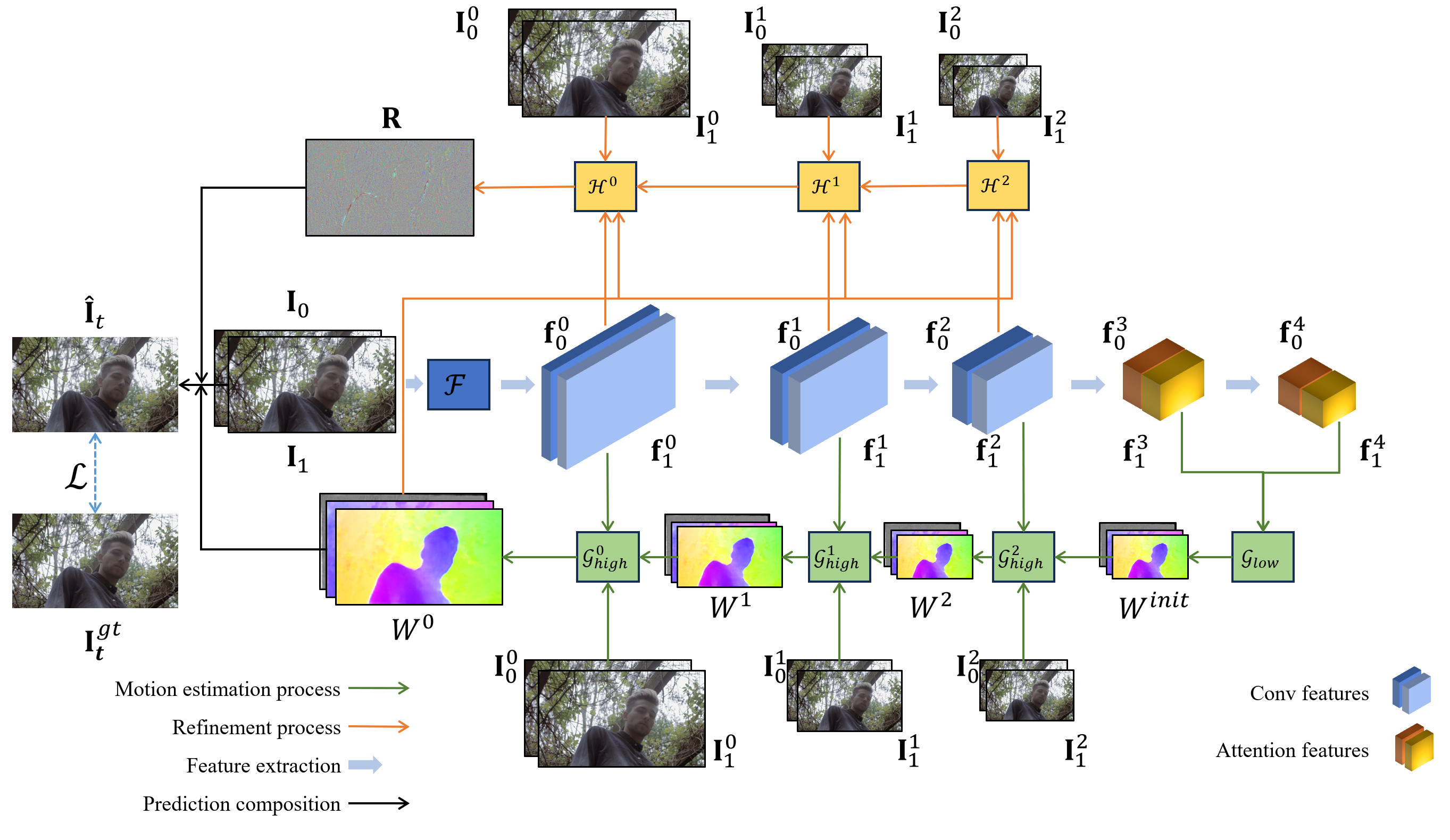} 
\caption{Illustration of our pipeline. Different processes are represented by colored lines. The images are first fed into the feature extractor $\mathcal{F}$ to produce five levels of features, where the first three levels are convolution-based features and the last two are attention-based features. These features are further processed by the flow estimator that consists of a low-res decoder $\mathcal{G}_{low}$ and three high-res decoders $\mathcal{G}_{high}^l$. This process, colored green, generates the motion flows and composition map, $W^0$. The high-res features, the input images and the estimated motion are further processed by the refinement module to generate the residual term $\textbf{R}$, colored orange. The final prediction, indicated by black lines, is the composition of the two input images plus the residual term. }
\label{fig:pipeline}
\end{figure*}

Given two images ($\textbf{I}_0$, $\textbf{I}_1$) and a timestep $t$ ($0 \le t \le 1$), the task is to synthesize an intermediate image as:
\begin{equation}
    \hat{\textbf{I}}_t=\mathcal{M}(\textbf{I}_0, \textbf{I}_1),
\end{equation}

where $\mathcal{M}$ is our model. Since predicting the middle frame is the most common usage case, we focus on $t=0.5$ here. 

Prediction of the intermediate frame is widely formulated as the composition of the two given images warped by motion flows plus a residual part, which is defined as:

\begin{equation}
    \hat{\textbf{I}}_t = \textbf{m} \odot \overleftarrow{\mathcal{W}}(\textbf{I}_0, \textbf{F}_{t \to 0} ) + (1-\textbf{m}) \odot \overleftarrow{\mathcal{W}}(\textbf{I}_1, \textbf{F}_{t \to 1} ) + \textbf{R},
    \label{eq:whole}
\end{equation}

where $\textbf{m}$ is a composition mask; $\odot$ denotes Hadamard product; $\overleftarrow{\mathcal{W}}$ is the backward warping operation; $\textbf{F}_{t \to 0}$ and $\textbf{F}_{t \to 1}$ represent motion flows between the intermediate frame and the given images; $\textbf{R}$ is the residual part.

Our model $\mathcal{M}$ is composed of a feature extractor $\mathcal{F}$, a flow estimator $\mathcal{G}$, and a refinement module $\mathcal{H}$. Figure \ref{fig:pipeline} illustrates the overall pipeline of our method. The input images are first fed to the feature extractor $\mathcal{F}$ to obtain features at five levels. The features are further passed to multiple decoders of the flow estimator $\mathcal{G}$ to estimate the motion flows and the mask. The residual part is then produced by $\mathcal{H}$ using the input images, image features and the estimated flow information. Eventually we obtain the final prediction of the frame $\hat{\textbf{I}}_t$ by composition. Next we elaborate each part in detail.

\textbf{Feature Extractor.}  The role of the feature extractor is to produce high quality features from the raw images, which are used for motion estimation and also image refinement. Pure convolution-based structures are a common choice \cite{rife, adacof, ifrnet}, and there are also approaches attempting to incorporate transformer-like structures where the attention mechanism enables long-range interactions within the frame or across frames \cite{vfiformer, ema}. We adopt a light-weight attention-based structure from \cite{ema}. For simplicity, we only borrow the appearance feature extraction part with inter-frame attention and the motion features are not incorporated.

More specifically, for the images with dimensions of ($H \times W$), we extract five levels of features:
\begin{equation}
    \{{\textbf{f}_0^l}, \textbf{f}_1^l, l \in [0, 4]\} = \mathcal{F}(\textbf{I}_0, \textbf{I}_1),
\end{equation}
where each level of features has spatial dimensions of ($\frac{H}{2^l}$, $\frac{W}{2^l}$). The first three levels of features are produced by convolution blocks and the last two levels are processed by the inter-frame attention blocks.



\textbf{Flow Estimator.} The flow estimator produces the composition mask $\textbf{m}$ and the motion flows, $\textbf{F}_{t \to 0}$ and $\textbf{F}_{t \to 1}$. However, directly computing the full resolution flows is challenging due to complex motions and diversity of scenes, therefore flows are usually estimated progressively in a coarse-to-fine fashion \cite{film, rife, ifrnet}. In this process, a sufficient big FoV is critical to encode rich context information and capture complex motions. For low-resolution features, stacking up multiple $3 \times 3$ convolutions might suffice, but for high-resolution features, it would be not sufficient. Simply enlarging the kernel size or adding more convolutions might not be an efficient solution for its largely increased computational cost. We propose to incorporate large-kernel depth-wise convolution to enable large FoV and at the same time reduce overhead. The idea of using large kernels has been investigated in other tasks such as image segmentation, detection and classification\cite{largekernel, replknet} , but has rarely been studied for VFI, which motivates us to exploit the effectiveness. Depth-wise convolution has advantages in this specific case. For feature maps with 64 channels, without changing the channel number, compared with a normal $3 \times 3$ convolution layer, a depth-wise convolution layer with kernel size $k=15$ consumes only half of FLOPs and uses half of parameters, but greatly enlarges FoV. Please also refer to the ablation experiments for more details.

Back to the design of our flow estimator, instead of using a uniform decoder block across all levels of features, we introduce a novel design featuring two types of decoders, low-res decoder $\mathcal{G}_{low}$ and high-res decoder $\mathcal{G}_{high}$, for efficient flow estimation. For low-resolution features, the low-res decoder, equipped with normal $3 \times 3$ convolution, is sufficiently effective, whereas for high-resolution features, the high-res decoders benefit from large-kernel depth-wise convolutions.

More specifically, the low-res decoder receives features, $\{\textbf{f}_0^4, \textbf{f}_1^4, \textbf{f}_0^3, \textbf{f}_1^3\}$, as inputs and generate a good initial estimation of the flows and composition mask, defined as:

\begin{equation}
    \{\textbf{F}_{t \to 0}^{init},  \textbf{F}_{t \to 1}^{init}, \textbf{m}^{init}\} = \mathcal{G}_{low}(\textbf{f}_0^4, \textbf{f}_1^4, \textbf{f}_0^3, \textbf{f}_1^3)
\end{equation}

Once the initial estimation is generated, we utilize three high-res decoders, $\mathcal{G}_{high}^l$ with $l\in[0, 1, 2]$, to progressively refine the results, which is defined as:

\begin{equation}
\begin{split}
    \Delta\textbf{W}^l  = \mathcal{G}_{high}^l & (\overleftarrow{\mathcal{W}}(\textbf{f}_0^l, \textbf{F}_{t \to 0}^{l+1}), \overleftarrow{\mathcal{W}}(\textbf{f}_1^l, \textbf{F}_{t \to 1}^{l+1}), \textbf{I}_0^l, \textbf{I}_1^l, \\
    & \overleftarrow{\mathcal{W}}(\textbf{I}_0^l, \textbf{F}_{t \to 0}^{l+1}), \overleftarrow{\mathcal{W}}(\textbf{I}_1^l, \textbf{F}_{t \to 1}^{l+1}), \textbf{W}^{l+1}) 
\end{split}
\end{equation}

\begin{equation}
    \textbf{W}^l = (\Delta\textbf{W}^l+ \textbf{W}^{l+1})_{\times 2} =\{\textbf{F}_{t \to 0}^{l},  \textbf{F}_{t \to 1}^{l}, \textbf{m}^{l}\}, 
\end{equation}

where $\textbf{W}^l$ represents the warping information, including motion flows and composition map; $(\cdot  )_{\times 2}$ is a bi-linear up-sampler; $\textbf{I}_1^l$ and $\textbf{I}_0^l$ are images down-sampled to level $l$. $\textbf{W}^0$ is the final output, and the bi-linear up-sampler is not applied.

For flow estimation, there are several design options in terms of the inputs. In \cite{film}, the decoder uses both the original image features and the warped features, where it requires a large channel width. In \cite{rife}, the flow estimation is based on a so-called IFBlock, whose inputs are the down-scaled images, the warped images, and the flow information. Its input only takes 17 channels, which is a light-weight design. However it relies only on the raw image information and might not benefit from the high quality features provided by the extractor. Our design takes the advantage of both the rich features and raw image information.

\textbf{Refinement Module.}

The composition of the warped input images provides an initial prediction of the intermediate frame. However, the prediction highly relies on the given pixel information and motion quality. If the model fails to find proper information due to motion errors or occlusions, it will produce unsatisfactory results. To overcome this limitation, a refinement module is usually adopted to compute a residual $\textbf{R}$ that compensates for the errors, as stated in Equation \ref{eq:whole}.

A common design for this refinement module is a UNet \cite{unet} structure that has an encoder with down-sampling blocks and a decoder with up-sampling blocks \cite{rife, ema, vfiformer}. In \cite{abme}, a GridNet \cite{gridnet} is used to perform refinement, which not only contains down-sampling and up-sampling blocks but also employs lateral blocks. We find this redundant in our framework since the feature extractor has already encoded frames into high-quality features on different levels, which make it unnecessary to employ another encoder in the refinement module. Therefore, we simplify this process by adopting a decoder-only structure, leading to effective and efficient refinement.


There are only three blocks in the refinement module $\mathcal{H}$ corresponding to high resolution features $\{{\textbf{f}_0^l}, \textbf{f}_1^l, l \in [0, 2]\}$. We find the features with inter-frame interactions $\{{\textbf{f}_0^l}, \textbf{f}_1^l, l \in [3, 4]\}$ do not provide boost in this stage, so we end up with this 3-level light-weight structure. Please refer to the ablation study where we compare different configurations of the refinement module.

For each level, it is defined as:

\begin{equation}
    \textbf{O}^l = \mathcal{H}^l(\overleftarrow{\mathcal{W}}(\textbf{f}_0^l), \overleftarrow{\mathcal{W}}(\textbf{f}_1^l), \textbf{I}_0^l, \textbf{I}_1^l, \overleftarrow{\mathcal{W}}(\textbf{I}_0^l), \overleftarrow{\mathcal{W}}(\textbf{I}_1^l), \widetilde{\textbf{W}}^l, \textbf{O}^{l+1}), 
\end{equation}

where $\textbf{I}_0^l$ and $\textbf{I}_1^l$ are scaled images, $\overleftarrow{\mathcal{W}}(\cdot)$ is a simplified representation for warping the image or features using $\mathbf{W}^0$ down-scaled to the corresponding size. $\textbf{O}^l$ is the output for level $l$. For $l=2$, $\textbf{O}^{l+1}$ is “None”, and for $l=0$, $\textbf{O}^l$ is the final residual $\textbf{R}$.

\textbf{HD-aware Augmentation}

A typical evaluation routine for VFI methods is training on Vimeo 90K dataset \cite{vimeo}, and evaluating on various datasets. We unavoidably face domain gaps between different sources of data. One of the issues is image resolution. Vimeo 90K dataset contains images with fixed size of $448 \times 256$, while datasets such as Xiph and SNU-FILM \cite{xiph, cain} feature HD images with size up to $1280 \times 720$ or $2048 \times 1080$. Generally applying the model directly on full resolution HD images would not give satisfactory results since objects and motions are scaled up significantly compared to the SD images.

A widely used strategy is to use the down-scaled images to estimate motion flows and scale up the flows back to the original resolution and perform image synthesis \cite{ifrnet, amt, ema, vfiformer}. However, we find this post-training method not working consistently well, e.g. performance increasing on SD datasets does not lead to corresponding improvement on HD datasets. In order to further stabilize the model performance on HD images, we incorporate this "synthesizing image with low-res flows" mechanism into the training process, enabling a more consistent ability to synthesize frames with both original and low-res flows, which we call HD-aware augmentation.

More specifically, after the standard training process, we further fine-tune the model by 5 epochs, where we randomly make the model utilize either the original flows estimated from the original size, or the down-scaled flows estimated from the down-sampled images. For simplicity, we only consider down-sampling scale of 0.5.

\subsection{Training Objectives}

Our model is trained solely on ground-truth frames without requiring any other pre-trained networks or extra supervisions. Our model is supervised by three terms.

\begin{equation}
    \mathcal{L} = \lambda_{ch}\mathcal{L}_{ch} + \lambda_{lap}\mathcal{L}_{lap} + \lambda_f\mathcal{L}_f
\end{equation}

The first term is a image reconstruction loss defined as: 

\begin{equation}
    \mathcal{L}_{ch} = \sqrt{(\hat{\textbf{I}}_t - \textbf{I}^{gt}_t)^2 + \epsilon^2},
\end{equation}

where $\epsilon$ is a small constant set to $1e^{-6}$. It is called Charbonnier penalty function \cite{Charbonnier} that servers as a  better reconstruction loss than the pure L1 loss.

The second term is a Laplacian pyramid loss \cite{lap} defined as:

\begin{equation}
    \mathbf{L}_{lap} = \sum_{i=1}^52^{i-1}\left\| L^i(\hat{\textbf{I}}_t) - L^i(\textbf{I}_t^{gt}) \right\|_1
\end{equation}

The $\mathbf{L}_{lap}$ is capable of retaining fine details, but focusing less on the low-frequency content such as color information. Therefore $\mathcal{L}_{ch}$ combined with $\mathbf{L}_{lap}$ provides better results.

We also add another frequency loss term $\mathcal{L}_f$, which is applied to image super resolution task \cite{frenq}. The loss is defined as:

\begin{equation}
      \mathcal{L}_{f} =  \frac{1}{2}\left\| \mathcal{T}_{\textit{fft}}^{\left| \cdot \right|}(\hat{\textbf{I}}_t) - \mathcal{T}_{\textit{fft}}^{\left| \cdot \right|}(\textbf{I}_t^{gt}) \right\|_1 + \frac{1}{2} \left\| \mathcal{T}_{\textit{fft}}^{\angle }(\hat{\textbf{I}}_t) - \mathcal{T}_{\textit{fft}}^{\angle }(\textbf{I}_t^{gt}) \right\|_1 ,
\end{equation}

where $\mathcal{T}_{\textit{fft}}^{\left| \cdot \right|}(\cdot)$ and $\mathcal{T}_{\textit{fft}}^{\angle }(\cdot)$ are amplitude and phase of the frequency components calculated by fast Fourier transform (FFT). This loss supervises the prediction in the frequency domain, providing supervision from a global holistic perspective in addition to the local spatial information.

\section{Experiments}

\begin{table*}[t]
\centering


\resizebox{\textwidth}{!}{%
\ra{1.3}
\begin{tabular}{lllllllllll}
\toprule
\multicolumn{1}{c}{\multirow{2}{*}{Method}} & \multicolumn{1}{c}{\multirow{2}{*}{Vimeo90k}} & \multicolumn{1}{c}{\multirow{2}{*}{UCF101}} & \multicolumn{4}{c}{SNU-FILM}                                                                                   & \multicolumn{2}{c}{Xiph}                        & \multicolumn{1}{c}{FLOPs} & \multicolumn{1}{c}{Params} \\ \cmidrule{4-9}
\multicolumn{1}{c}{}                        & \multicolumn{1}{c}{}                          & \multicolumn{1}{c}{}                        & \multicolumn{1}{c}{Easy} & \multicolumn{1}{c}{Medium} & \multicolumn{1}{c}{Hard} & \multicolumn{1}{c}{Extreme} & \multicolumn{1}{c}{2K} & \multicolumn{1}{c}{4K} & \multicolumn{1}{c}{(T)}   & \multicolumn{1}{c}{(M)}    \\ \midrule

        M2M & 35.47/0.9778 & 35.28/0.9694 & 39.66/0.9904 & 35.74/0.9794 & 30.30/0.9360 & 25.08/0.8604 & 36.44/\textbf{0.943} & 33.92/0.899 & 0.26 & 7.6  \\ 
        IFRNet-S & 35.59/0.9786 & 35.28/0.9691 & 39.96/0.9905 & 35.92/0.9795 & 30.36/0.9357 & 25.05/0.8582 & 35.87/0.936 & 33.80/0.891 & 0.12 & 2.8  \\ 
        RIFE & 35.61/0.9779 & 35.28/0.9690 & 39.80/0.9903 & 35.76/0.9787 & 30.36/0.9351 & 25.27/0.8601 & 36.19/0.938 & 33.76/0.894 & 0.20 & 9.8  \\ 
        IFRNet & 35.80/0.9794 & 35.29/0.9693 & \underline{40.03}/0.9905 & 35.94/0.9793 & 30.41/0.9358 & 25.05/0.8587 & 36.00/0.936 & 33.99/0.893 & 0.21 & 5.0  \\ 
        AMT-S \dag & 35.97/\underline{0.9800} & \underline{35.35}/\underline{0.9697} & 39.95/0.9905 & \underline{35.98}/\underline{0.9796} & 30.60/\underline{0.9369} & 25.30/0.8625 & 36.14/0.940 & \underline{34.32}/\underline{0.902} & 0.12 & 3.0  \\ 
        EMA-S & \underline{36.07}/0.9797 & 35.34/0.9696 & 39.81/\underline{0.9906} & 35.88/0.9795 & \underline{30.69}/\textbf{0.9375} & \underline{25.47}/\underline{0.8632} & \underline{36.55}/\underline{0.942} & 34.25/\underline{0.902} & 0.21 & 14.5  \\ 
        Ours & \textbf{36.24}/\textbf{0.9804} & \textbf{35.40}/\textbf{0.9698} & \textbf{40.16}/\textbf{0.9908} & \textbf{36.14}/\textbf{0.9799} & \textbf{30.77}/\textbf{0.9375} & \textbf{25.54}/\textbf{0.8633} & \textbf{36.60}/\textbf{0.943} & \textbf{34.35}/\textbf{0.903} & 0.14 & 3.1  \\ 

        \midrule
        ToFLow & 33.73/0.9682 & 34.58/0.9667 & 39.08/0.9890 & 34.39/0.9740 & 28.44/0.9180 & 23.39/0.8310 & 33.93/0.922 & 30.74/0.856 & 0.62 & 1.4  \\ 
        SepConv & 33.79/0.9702 & 34.78/0.9669 & 39.41/0.9900 & 34.97/0.9762 & 29.36/0.9253 & 24.31/0.8448 & 34.77/0.929 & 32.06/0.880 & 0.38 & 21.7  \\ 
        AdaCoF & 34.47/0.9730 & 34.90/0.9680 & 39.80/0.9900 & 35.05/0.9754 & 29.46/0.9244 & 24.31/0.8439 & 34.86/0.928 & 31.68/0.870 & 0.36 & 21.8  \\ 
        CAIN & 34.65/0.9730 & 34.91/0.9690 & 39.89/0.9900 & 35.61/0.9776 & 29.90/0.9292 & 24.78/0.8507 & 35.21/0.937 & 32.56/0.901 & 1.29 & 42.8  \\ 
        DAIN & 34.71/0.9756 & 34.99/0.9683 & 39.73/0.9902 & 35.46/0.9780 & 30.17/0.9335 & 25.09/0.8584 & 35.95/0.940 & 33.49/0.895 & 5.51 & 24.0  \\ 
        
        BMBC & 35.01/0.9764 & 35.15/0.9689 & 39.90/0.9902 & 35.31/0.9774 & 29.33/0.9270 & 23.92/0.8432 & 32.82/0.928 & 31.19/0.880 & 2.50 & 11.0  \\ 
        SoftSplat & 36.10/0.9802 & 35.39/0.9697 & 39.88/0.9897 & 35.68/0.9772 & 30.19/0.9312 & 24.83/0.8500 & 36.62/\underline{0.944} & 33.60/0.901 & 1.00 & 7.7  \\ 
        ABME & 36.18/0.9805 & 35.38/0.9698 & 39.59/0.9901 & 35.77/0.9789 & 30.58/0.9364 & 25.42/0.8639 & 36.53/\underline{0.944} & 33.73/0.901 & 1.30 & 18.1  \\ 
        IFRNet-L & 36.20/0.9808 & 35.42/0.9698 & 40.10/0.9906 & \underline{36.12}/0.9797 & 30.63/0.9368 & 25.27/0.8609 & 36.21/0.937 & 34.25/0.895 & 0.79 & 19.7  \\ 
        AMT-L \dag & 36.35/0.9813 & 35.42/0.9699 & 39.95/0.9904 & 36.09/0.9797 & 30.75/0.9377 & 25.41/0.8633 & 36.31/0.941 & 34.51/0.904 & 0.58 & 12.9  \\ 
        EMA-L * & 36.50/0.9800 & 35.42/\underline{0.9700} & 39.58/0.9890 & 35.86/0.9790 & 30.80/0.9380 & \underline{25.59}/0.8640 & 36.74/\underline{0.944} & 34.55/0.906 & 0.91 & 65.6  \\ 
        VFIformer & 36.50/0.9816 & 35.43/\underline{0.9700} & \underline{40.13}/\underline{0.9907} & 36.09/\underline{0.9799} & 30.67/0.9378 & 25.43/\underline{0.8643} & \underline{36.82}/\textbf{0.946} & 34.43/\underline{0.907} & 47.71 & 24.1  \\

        AMT-G \dag & \underline{36.53}/\textbf{0.9819} & \textbf{35.45}/\underline{0.9700} & 39.88/0.9905 & \underline{36.12}/0.9798 & \underline{30.78}/\underline{0.9379} & 25.43/0.8640 & 36.42/0.942 & \underline{34.65}/0.905 & 2.07 & 30.6  \\  
        Ours-L & \textbf{36.65}/\underline{0.9818} & \underline{35.44}/\textbf{0.9702} & \textbf{40.17}/\textbf{0.9909} & \textbf{36.22}/\textbf{0.9803} & \textbf{30.87}/\textbf{0.9384} & \textbf{25.71}/\textbf{0.8658} & \textbf{36.89}/\textbf{0.946} & \textbf{34.88}/\textbf{0.908} & 0.61 & 20.0  \\

\bottomrule
\end{tabular}
}

\caption{Quantitative comparison between VFI methods. We divide the methods into two groups based on the FLOPs and model size. For each group, the best result is shown in \textbf{bold} and the second best is \underline{underlined}. * indicates the result without test-time augmentation for fair comparison. \dag represents our duplicated results using the official code.}
\label{tab: overall comparison}
\end{table*}

\newcommand{\firstcwidth}{0.165}
\newcommand{\othercwidth}{0.0835}
\newcommand{\colmspace}{0.1em}

\begin{figure*}[t]
	\centering
	\resizebox{0.91\textwidth}{!}{
		\begin{tabular}{c@{\hspace{\colmspace}}c@{\hspace{\colmspace}}c@{\hspace{\colmspace}}c@{\hspace{\colmspace}}c@{\hspace{\colmspace}}c@{\hspace{\colmspace}}c@{\hspace{\colmspace}}c@{\hspace{\colmspace}}c@{\hspace{\colmspace}}c@{\hspace{\colmspace}}c}

                \includegraphics[width=\firstcwidth\textwidth]{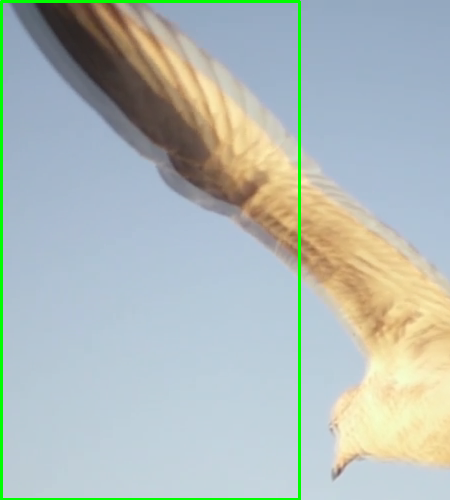} &
			\includegraphics[width=\othercwidth\textwidth]{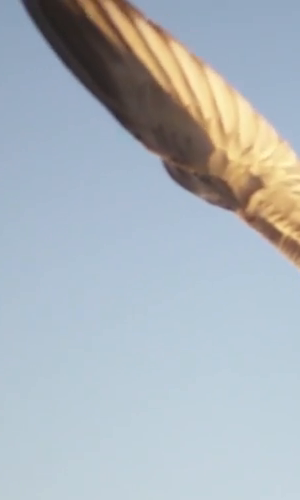} &
			\includegraphics[width=\othercwidth\textwidth]{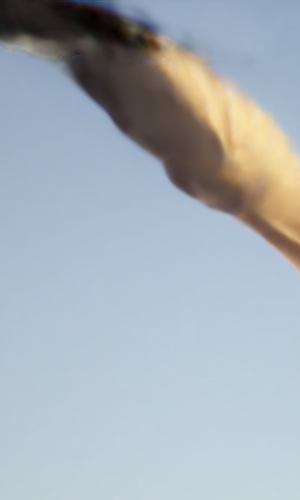} &
                    \includegraphics[width=\othercwidth\textwidth]{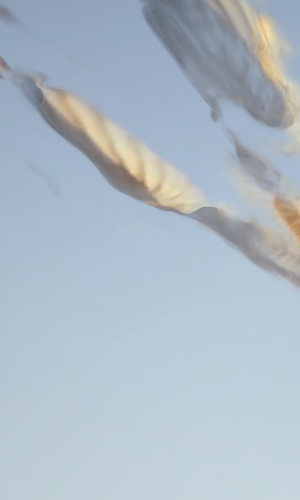} &
                    \includegraphics[width=\othercwidth\textwidth]{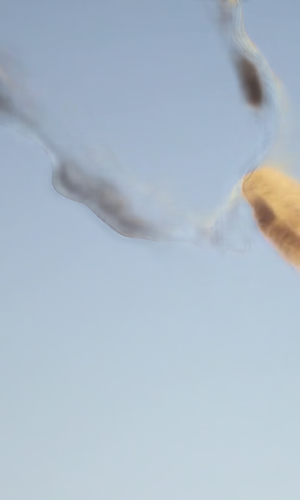} &
                    \includegraphics[width=\othercwidth\textwidth]{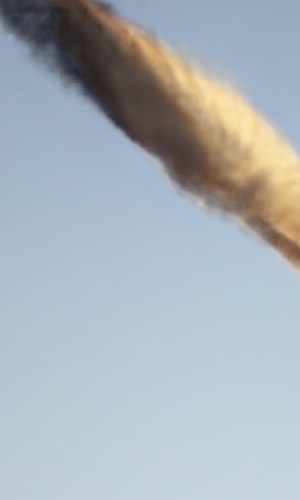} &
                    \includegraphics[width=\othercwidth\textwidth]{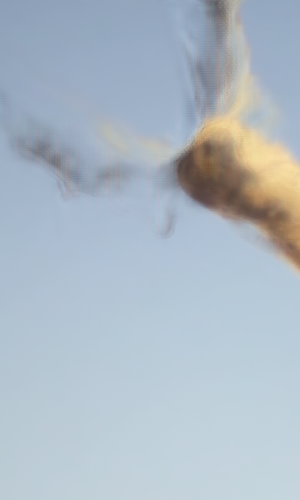} &
                    \includegraphics[width=\othercwidth\textwidth]{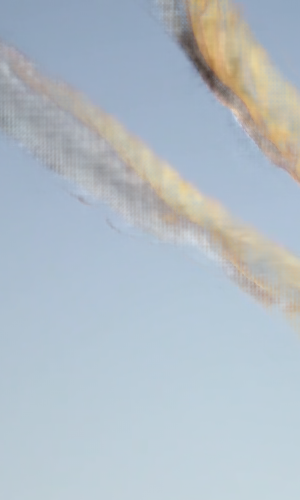} &
                    \includegraphics[width=\othercwidth\textwidth]{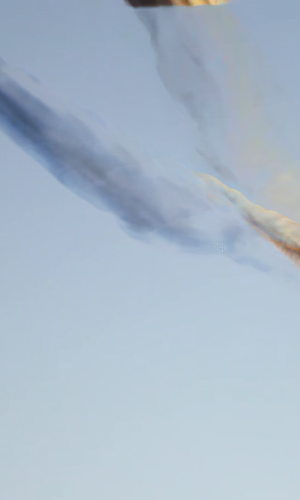} &
                    \includegraphics[width=\othercwidth\textwidth]{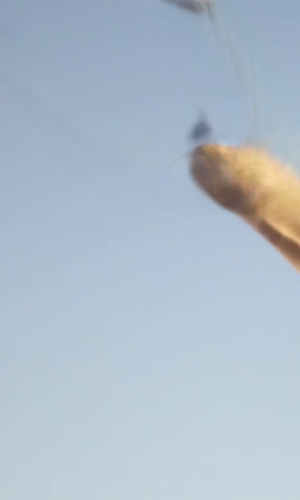} &
                    
			\includegraphics[width=\othercwidth\textwidth]{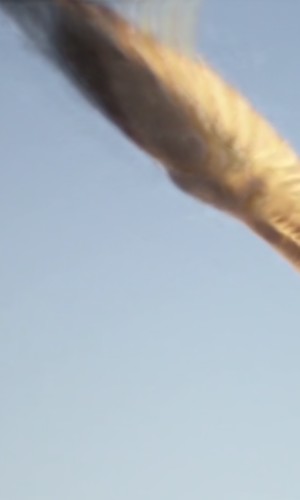}  \\
			\includegraphics[width=\firstcwidth\textwidth]{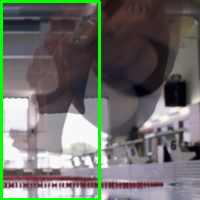} &
			\includegraphics[width=\othercwidth\textwidth]{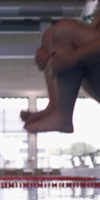} &
			\includegraphics[width=\othercwidth\textwidth]{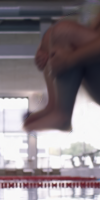} &
                    \includegraphics[width=\othercwidth\textwidth]{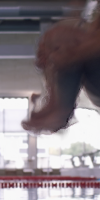} &
                    \includegraphics[width=\othercwidth\textwidth]{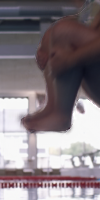} &
                    \includegraphics[width=\othercwidth\textwidth]{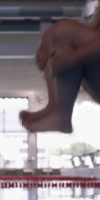} &
                    \includegraphics[width=\othercwidth\textwidth]{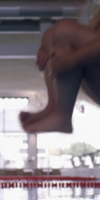} &
                    \includegraphics[width=\othercwidth\textwidth]{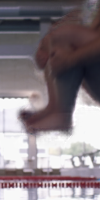} &
                    \includegraphics[width=\othercwidth\textwidth]{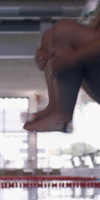} &
                    \includegraphics[width=\othercwidth\textwidth]{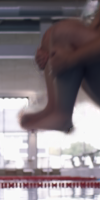} &
                    
			\includegraphics[width=\othercwidth\textwidth]{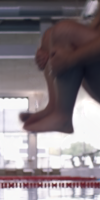}  \\

                \includegraphics[width=\firstcwidth\textwidth]{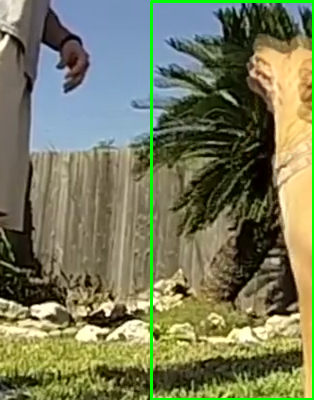} &
			\includegraphics[width=\othercwidth\textwidth]{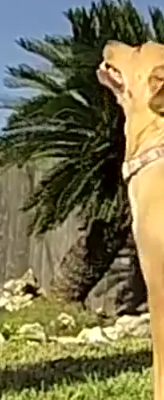} &
			\includegraphics[width=\othercwidth\textwidth]{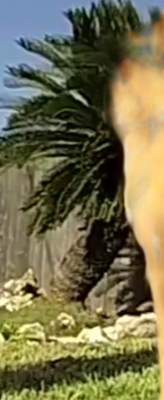} &
                    \includegraphics[width=\othercwidth\textwidth]{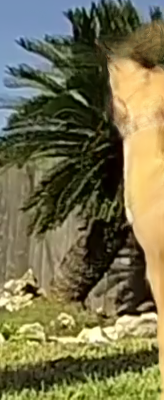} &
                    \includegraphics[width=\othercwidth\textwidth]{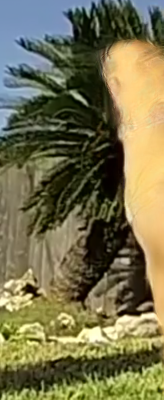} &
                    \includegraphics[width=\othercwidth\textwidth]{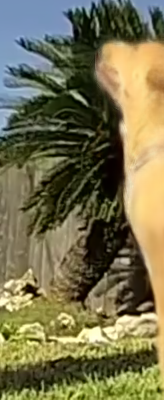} &
                    \includegraphics[width=\othercwidth\textwidth]{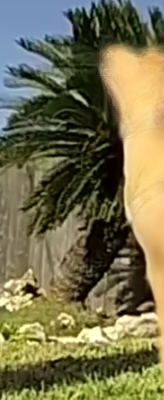} &
                    \includegraphics[width=\othercwidth\textwidth]{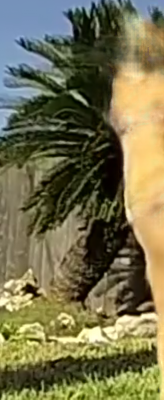} &
                    \includegraphics[width=\othercwidth\textwidth]{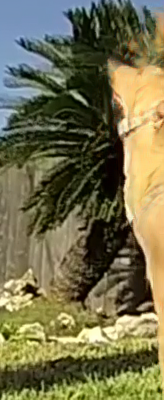} &
                    \includegraphics[width=\othercwidth\textwidth]{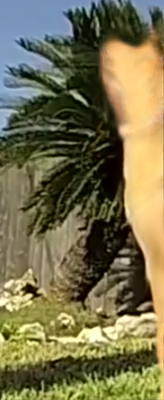} &
                    
			\includegraphics[width=\othercwidth\textwidth]{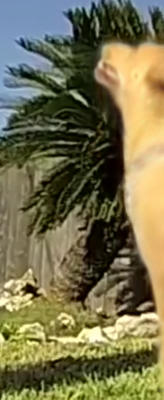}  \\

               \small Input Overlay & GT & ABME & AdaCof & IFRNet & EMA & AMT & RIFE & SoftSplat & VFIformer  & Ours \\ 
			
	\end{tabular}}

	\caption{Qualitative comparison.}
	\label{fig:prediction_results}
\end{figure*}

\begin{table}[t]
\centering

\ra{1.3}
\resizebox{0.5\linewidth}{!}{
\begin{tabular}{ccccc}
\toprule
$\mathcal{L}_{ch}$     & $\mathcal{L}_{lap}$      & $\mathcal{L}_f$      & & \multicolumn{1}{c}{PSNR (dB)} \\ \hline
  \ding{52}     &  \ding{56}      &   \ding{56}    & &   36.1   \\ 
\ding{56}     &  \ding{52}      &   \ding{56}    & & 15.04 \\
\ding{52}     &  \ding{52}      &   \ding{56}    & & 36.17\\
\ding{52}     &  \ding{52}      &   \ding{52}    & & \textbf{36.24}\\

\bottomrule
\end{tabular}}

\caption{Ablation study on different losses.}
\label{tab:loss}
\end{table}

\begin{table}[t]

\centering
\resizebox{0.95\linewidth}{!}{
\ra{1.3}
\begin{tabular}{lllccc}
\toprule
\multicolumn{3}{l}{Block conf. in $\mathcal{G}_{high}^l$} & PSNR (dB)  & FLOPs (G)   & Params (M)\\ 
\hline
  \multicolumn{3}{l}{Conv k=[3, 3, 3] } &   36.11  & 150.74 & 3.21\\ 
  \multicolumn{3}{l}{DSConv k=[5, 5, 5]}      &   36.12 & 127.67 & 3.03  \\
  \multicolumn{3}{l}{DSConv k=[7, 7, 7] }    &   36.13  & 130.65 & 3.05\\
  \multicolumn{3}{l}{DSConv k=[7, 15, 15] }  &   \textbf{36.24}  & 139.73 & 3.08\\

\bottomrule
\end{tabular}
}

\caption{Ablation study on high-res decoder settings.}
\label{tab:blocks}
\end{table}

\begin{table}[t]

\centering
\resizebox{0.9\linewidth}{!}{
\ra{1.3}
\begin{tabular}{lllccc}
\toprule
\multicolumn{3}{l}{$\mathcal{H}$ conf.} & PSNR (dB)  & FLOPs (G)   & Params (M)\\ 
\hline
  \multicolumn{3}{l}{UNet} &   36.21  & 207.27 & 14.49\\ 
  \multicolumn{3}{l}{5L decoder-only}      &   36.20 & 179.57 & 9.49  \\
  \multicolumn{3}{l}{4L decoder-only}    &   36.22  & 159.68 & 4.37\\
  \multicolumn{3}{l}{3L decoder-only}  &   \textbf{36.24}  & 139.73 & 3.08\\
  \multicolumn{3}{l}{2L decoder-only}  & 36.13  & 119.63 & 2.76\\

\bottomrule
\end{tabular}
}

\caption{Results of different structures of refinement module.}
\label{tab:refinemodule_config}
\end{table}

\begin{table}[t]

\centering
\ra{1.3}
\resizebox{0.7\linewidth}{!}{
\begin{tabular}{lllc}
\toprule
\multicolumn{3}{l}{Training config in $\mathcal{G}_{high}^l$} & PSNR (dB)   \\ 
\hline
  \multicolumn{3}{l}{One-stage training} &   36.01   \\ 
  \multicolumn{3}{l}{One-stage training with extra sup.}      &   36.05   \\
  \multicolumn{3}{l}{Two-stage training}    &   \textbf{36.24}   \\

\bottomrule
\end{tabular}}

\caption{Ablation study on pre-training.}
\label{tab:pretrain}
\end{table}

\begin{table}[t]

\centering
\ra{1.3}
\resizebox{0.85\linewidth}{!}{
\begin{tabular}{lcccc}
\toprule
Method  & Easy & Medium & Hard & Extreme    \\ 
\hline
  With orig. flow  & 40.11 & 35.86 & 30.23 & 25.02 \\ 
 With ds. flow   & 39.86 & 35.98 & 30.71 & 25.5  \\
HD-aware aug      & \textbf{40.16} & \textbf{36.14} & \textbf{30.77} & \textbf{25.54}   \\

\bottomrule
\end{tabular}}

\caption{Ablation study on HD augmentation.}
\label{tab:hdaug}
\end{table}






\subsection{Implementation Details}
\textbf{Model Configuration.} We use $C$ to describe the base-width of feature channels. For the feature extractor, feature channels for each level are [$C$, $2C$, $4C$, $8C$, $16C$]. For the decoders, the low-res decoder has internal feature number of $2C$. The three high-res decoders work on feature channels of $4C$, $2C$ and $C$ respectively. For the image refinement module, the internal channel widths for the three blocks are $8C$, $4C$ and $2C$ respectively. We present two versions of our framework. For a light-weight one, we set $C=16$ and use two attention blocks for each level. For a relatively larger model, we have $C=32$ and assign four attention blocks.

\textbf{Datasets.} We conduct experiments on four widely used datasets, Vimeo90K \cite{vimeo}, UCF101 \cite{ucf}, SNU-FILM \cite{cain} and Xiph \cite{xiph}, covering SD and HD scenarios and rich types of scenes.

\textbf{Training details.} 
The model is trained solely on Vimeo-90 training set and evaluated on all four datasets. Images are randomly cropped to a patch of $256 \time 256$. In addition, we apply data augmentation methods: random vertical and horizontal flipping, random scale with [1, 2], random rotation with $[-45^\circ, 45^\circ]$ and triplet reversing. The model is optimized by AdamW \cite{adamw} with batch size of $32$, $\beta_1=0.9$, $\beta_2=0.999$, and weight decay of $1e-4$. The learning rate is decreased by the cosine annealing schedule from $2e-4$ to $2e-5$. 

We adopt a two-stage training scheme, training the flow estimator first and continuing training the whole framework by including the refinement module. Each stage is performed with the same routine. The only difference is that the first stage only considers composition of the two warped images, whereas the second stage involves the residual prediction.

\subsection{Comparisons with Other Methods}

We use PSNR and SSIM to perform quantitative comparison between different methods, including M2M \cite{m2m}, IFRNet \cite{ifrnet}, RIFE \cite{rife}, AMT \cite{amt}, EMA \cite{ema}, ToFlow\cite{toflow}, SepConv \cite{sepconv}, AdaCof \cite{adacof}, CAIN \cite{cain}, DAIN \cite{dain}, BMBC \cite{bmbc}, SoftSplat \cite{softsplat}, ABME \cite{abme}, and VFIformer \cite{vfiformer}. 

Table \ref{tab: overall comparison} shows the experimental results and comparison between methods. We divide the models into two groups based on computational overhead (FLOPs) and number of parameters. For the light-weight models, ours obtains the best PSNR for all the entries. Besides, compared with the overall second best method, EMA-S, our model requires $33\%$ less FLOPs and $79\%$ less parameters. For large models, we also achieve clear improvement over the previous best models for almost all the entries except for SSIM on Vimeo90k and PSNR on UCF101, where ours is very close to the best one, differing only in the last decimal digit. Additionally, our large model again shows clear advantage in terms of FLOPs and model size, requiring $70\%$ and $35\%$ less FLOPs and parameters than the second best, AMT-G. Figure \ref{fig:prediction_results} illustrates some visual results for a qualitative comparison.

\subsection{Ablation Study}
We further conduct experiments with different configurations to verify effectiveness of the components, which include loss functions, block configurations for the flow estimator, structures of the refinement module, pre-training and HD augmentation.

\textbf{Loss functions.} We study the effect of the three loss functions in the training. As described in Table \ref{tab:loss}, training the model purely with $\mathcal{L}_{ch}$ gives us a baseline score of 36.1 dB on PSNR. The model trained only with $\mathcal{L}_{lap}$ does not produce a plausible score, only 15.04dB, but interestingly the training loss decreases normally. The reason is that $\mathbf{L}_{lap}$ is capable of capturing high-frequency details but might lose some low-frequency information such as color. Therefore  $\mathcal{L}_{lap}$ and $\mathcal{L}_{ch}$ are a good combination for such case, reaching PSNR of 36.17dB. Furthermore, with the help of $\mathcal{L}_{f}$, the performance is further enhanced to 36.24 dB.

\textbf{Block configurations for high-res decoders.} We investigate different settings for the high-res decoders. As shown in Table \ref{tab:blocks}, using only three $3 \times 3$ convolution layers only achieves PSNR of 36.11dB. By replacing all the normal convolutions to the depth-wise separable convolutions, we get slightly better performance with reduced FLOPs (127.67 vs 150.74) and parameters (3.03 vs 3.21). Increasing all kernel sizes to 7 further slightly improves the score with limited overhead. Our final design adopts the setting of [7, 15, 15], which boosts the PSNR by 0.13dB compared with the normal-convolution version, while requires less FLOPs (7\% less) and parameters (4\% less) .

\textbf{Structures of refinement module.} To verify the effectiveness of our decoder-only refinement module, we conduct experiments on various structure settings, shown in Table \ref{tab:refinemodule_config}. XL means how many levels of features we feed to the refinement module. Compared with our optimal structure with 3 levels, the UNet structure does not give better performance while largely increasing FLOPs and model size. For decoder-only structures, using inter-frame attention features (4L and 5L structures) does not boost the performance and on the contrary, using only 2 levels (2L structure) of convolution features is clearly not sufficient. The ablation study indicates that fully utilizing all three convolution-based features maximizes the performance.

\textbf{Pre-training.} We adopt an effective two-stage training scheme in our framework, where the flow-estimator is firstly trained to predict plausible flow information and the refinement is further added for the second stage training. We find the two-stage training practically useful and we study different training strategies in Table \ref{tab:pretrain}. Directly training the whole framework from scratch only achieves PSNR of 36.01dB. By applying extra supervision to the intermediate flow estimation process, we obtain slightly better performance of 36.05dB. Taking the two-stage training boosts the performance to 36.24dB.

\textbf{HD-aware augmentation.} We also investigate effectiveness of our HD-aware augmentation strategy. As shown in Table \ref{tab:hdaug}, the model performance with the original flow (calculated by original inputs) is acceptable on the Easy set but deteriorates on the remaining three sets. In Contrast, the utilization of down-sampled flow (calculated by the down-sampled inputs) improves the performance on those three sets but fails to retain the score on the Easy set. Applying our HD-aware augmentation strategy yields consistent improvement over all four sets.

\section{Conclusion}
In this paper, we introduce an efficient framework for VFI with carefully designed components, which achieves State-of-the-art performance with much less FLOPs and parameters. More specifically, we first propose a flow estimator using depth-wise convolution with large kernels, enabling effective handling of large motions with reduced overhead. Furthermore, our proposed refinement module, featuring a decoder-only structure, provides an efficient process of refinement. Additionally, we introduce an HD-aware augmentation strategy to address "performance drop" issue on HD images. We hope this would establish a solid baseline for future research of the community.

\clearpage

\bibliography{aaai24}

\end{document}